\documentclass[conference]{IEEEtran}
\IEEEoverridecommandlockouts

\usepackage{cite}
\usepackage{multirow}
\usepackage{amsmath,amssymb,amsfonts}
\usepackage{algorithmic}
\usepackage{textcomp}
\usepackage{xcolor}
\usepackage{booktabs}
\usepackage{algorithm}
\usepackage{algorithmic}
\usepackage{adjustbox}
\usepackage{caption}
\usepackage{graphicx, subfig}
\usepackage{booktabs} 
\usepackage{bm}

\def\BibTeX{{\rm B\kern-.05em{\sc i\kern-.025em b}\kern-.08em
    T\kern-.1667em\lower.7ex\hbox{E}\kern-.125emX}}
\begin{document}

\title{Temporal and Spatial Online Integrated Calibration for Camera and LiDAR\\
\footnotesize \textsuperscript
{{}}
\thanks{This work was supported by the National High Technology Research and Development Program of China under Grant No. 2018YFE0204300, and the National Natural Science Foundation of China under Grant No. U1964203, and sponsored by Tsinghua University-Didi Joint Research Center for Future Mobility. 
 }
}

\author{Shouan Wang$^{1}$,  Xinyu Zhang$^{1}$$^,$$^*$, GuiPeng Zhang$^{1}$, Yijin Xiong$^{1}$, Ganglin Tian$^{1}$,\\ Shichun Guo$^{1}$, Jun Li$^{1}$, Pingping Lu$^{2}$, Junqing Wei$^{3}$ and Lei Tian$^{4}$

	\thanks{$^{1}$The authors are with the State Key Laboratory of Automotive Safety and Energy, the School of Vehicle and Mobility, Tsinghua University, 100084.}
	\thanks{$^{2}$Pingping Lu is with University of Michigan.}%
	\thanks{$^{3}$Junqing Wei is with DiDi Co. Ltd.}%
	\thanks{$^{4}$Lei Tian is with SINOTRUK Co. Ltd.}%
	\thanks{*Author to whom correspondence should be addressed. (e-mail:
	xyzhang@tsinghua.edu.cn)}%
}

\maketitle

\begin{abstract}
While camera and LiDAR are widely used in most of the assisted and autonomous driving systems, only a few works have been proposed to associate the temporal synchronization and extrinsic calibration for camera and LiDAR which are dedicated to online sensors data fusion. 
The temporal and spatial calibration technologies are facing the challenges of lack of relevance and real-time. In this paper, we introduce the pose estimation model and environmental robust line features extraction to improve the relevance of data fusion and instant online ability of correction. Dynamic targets eliminating aims to seek optimal policy considering the correspondence of point cloud matching between adjacent moments. The searching optimization process aims to provide accurate parameters with both computation accuracy and efficiency.
To demonstrate the benefits of this method, we evaluate it on the KITTI benchmark with ground truth value. 
In online experiments, our approach improves the accuracy by 38.5\% than the soft synchronization method in temporal calibration. While in spatial calibration, our approach automatically corrects disturbance errors within 0.4 second and achieves an accuracy of 0.3-degree. 
This work can promote the research and application of sensor fusion.

\end{abstract}


\section{INTRODUCTION}

\begin{figure}
	\includegraphics[width=1.0\columnwidth]{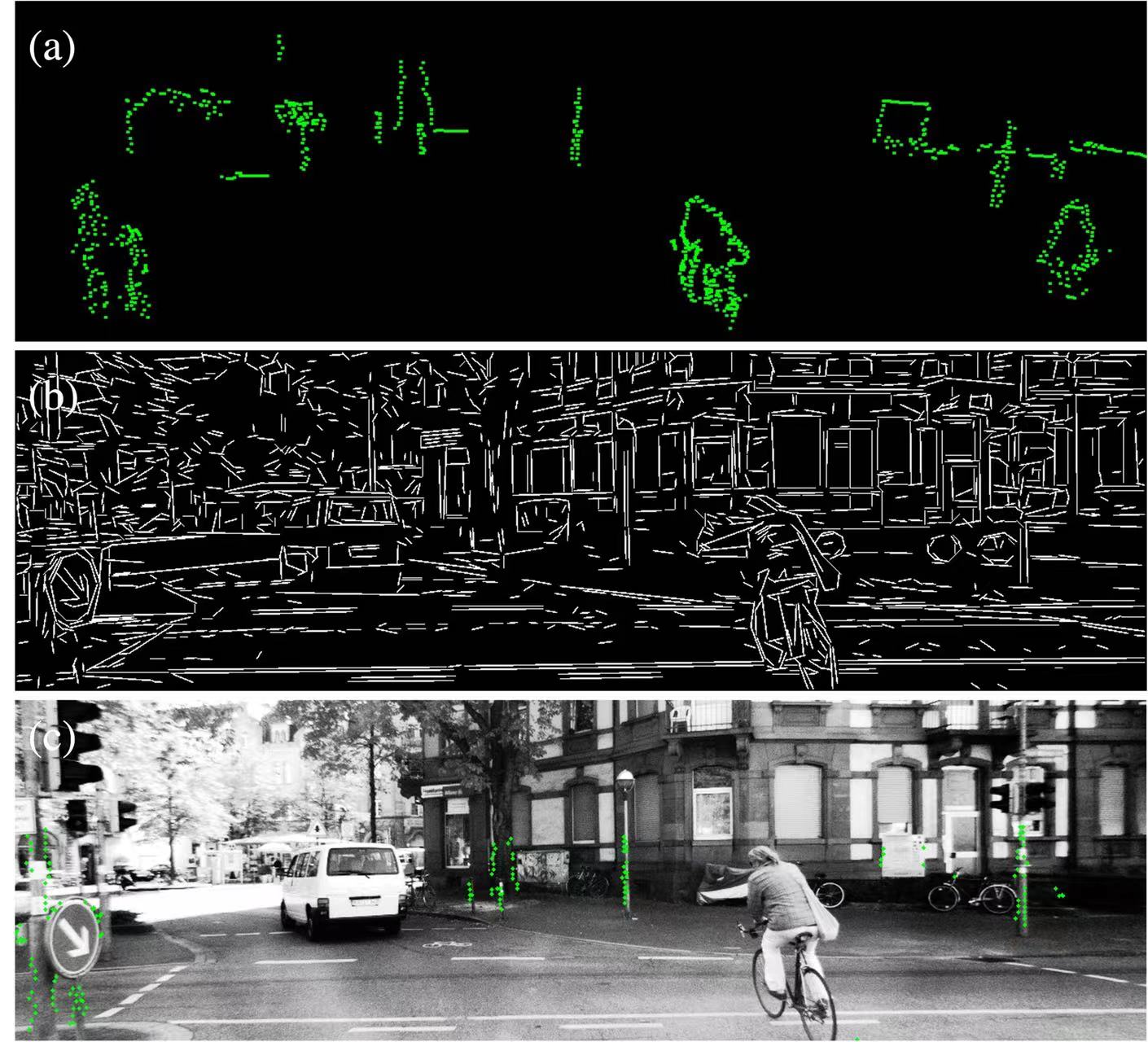}
	\caption{Line features are extracted from a point cloud (a) at moment $t+e_{t}$ and an image (b) at moment $t$ of the same scene. Through pose estimation model and calculation of correspondence between 2D and 3D line features, the projected point cloud line features (c) are well aligned with the image line features. }
	\label{main_first}
\end{figure}

There have been many studies on multi-sensor fusion in autonomous vehicles in recent years, benefiting from the multi-sensors equipped in autonomous vehicles. Sensor fusion is an essential and critical strategy to enhance the performance and ensure the reliability of perception modules in ADAS,
such as in target classification\cite{qi2017pointnet}, segmentation\cite{ronneberger2015u}, SLAM\cite{shan2020lio}, etc. Both temporal and spatial calibrations are the most critical procedures for sensor fusion. There have been a number of methods and systems for temporal calibrations and spatial calibrations.

For temporal calibration, hardware-based approaches are widely adopted for its accuracy. However, a unified clock signal and global timestamp are required and sensor hardware trigger interface must be reserved, which is not universally applicable to all devices; software-based approaches such as soft-synchronous interpolation\cite{wu2018soft}, values are aligned by curve fitting, which brings accuracy problems because the data are theoretical values obtained by fitting to the theoretical value. 

\begin{figure*}[]
	\centering
	\includegraphics[width=\textwidth]{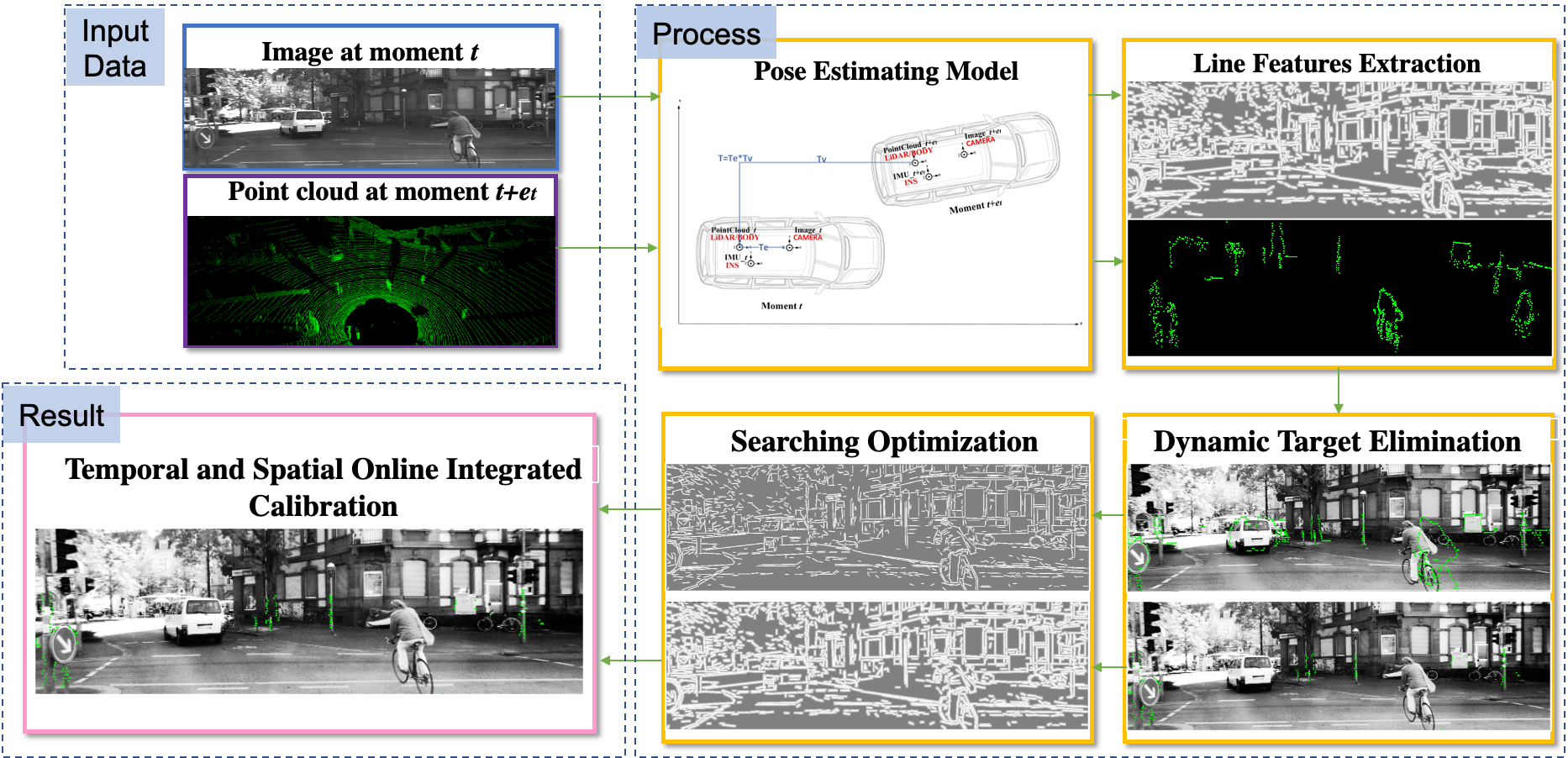}
	\caption{The architecture of the proposed calibration system.}
	\label{kuangjia_pic}
\end{figure*}

For spatial calibration, specially designed objects are required for traditional manual calibration approaches, such as manually selected points\cite{kato2015open}, which brings redundant manual operations. Long-time operation and varying loads can lead to slight drifts and deviations in extrinsic parameters. Current artificially designed targets\cite{geiger2012automatic} are utilized to calibrate the extrinsic parameters in automatic calibration works, but it's only at the stage of the laboratory. Some feature-based calibration method\cite{castorena2016autocalibration} utilizes edge features to compute the extrinsic parameters. However, these features are not well corresponded to each other under some scenarios.

The current temporal and spatial calibration modules are not well integrated and associated, and only a few works were proposed to solve the temporal and spatial calibration integration for camera and LiDAR through calibration boards while ignoring the online performance.

Therefore, in this paper we aim to combine the temporal-spatial calibration into online integrated calibration. For this purpose, sensors' pose estimation model between adjacent moments is introduced for online integrated calibration, and line features are used to correct the error of the pose estimation model.

The main contributions of this paper are as follows:

\text { 1) }A novel pose estimation model is proposed to decrease time delay and extrinsic offset in temporal and spatial online integrated calibration procedures.

\text { 2) }We introduce a method for eliminating dynamic point clouds which only used the association of adjacent two frames instead of detection boxes based on prior information

\text { 3) }We introduced a searching optimization method to improve the optimization efficiency and proposed a new set of accuracy assessment metrics to evaluate the accuracy of the calibration results.

\section{RELATED WORKS}
In this section, we discuss the development of temporal and spatial calibrations. 
Temporal calibration and spatial calibration are poorly coupled and generally separated. For the temporal calibration part, hardware-based methods have been widely applied, due to their exceptional accuracy. 
The proposal of TriggerSync\cite{english2015triggersync} represented the high accuracy of the hardware triggered method, however, it was low robust to the false correlation of trigger pulses due to unexpected delays, and more interfaces were required. 
To solve the problem of an unexpected delay, the Global Position System(GPS)\cite{dana1997global} signal was proposed as the timing signal. However, in the GPS rejected region, timing accuracy was seriously affected by the signal problem. To overcome this drawback, Marsel Faizullin et al.\cite{faizullin2021open} proposed a microcontroller-based platform for GPS signal emulation that can be extended to arbitrary multi-sensors beyond LiDAR-imu. Hannes Sommer et al.\cite{sommer2017low} used simple external devices, including LED (for camera) or photodiodes (for LiDAR), to achieve temporal calibration. 
This was done at the cost of additional equipment in exchange for a high rate and high precision estimate of time offset. 
Hardware-triggered interfaces or support from other hardware were required by all of the above approaches, which lead to their limited use in all scenarios. 
Instead, software-based methods were gradually popular because of its low hardware requirements. The message\_filter method was based on Robot Operating System(ROS), however, it only adopted a simple strategy to match timestamps when dealing with sensor data of different frequencies, which performed poorly in high-precision data fusion scenarios. 
bidirectional synchronization methods such as TICSync\cite{harrison2011ticsync}, PTP or NTP, performed well but require sensor firmware support, which was rare and expensive.

For the spatial calibration part, checkerboard and selected points were first chosen to solve the calibration problem between LiDAR and camera\cite{scaramuzza2007extrinsic}. The main limitation of the above approaches was manual and time-consuming. 
To overcome the above limitations, several ways were proposed by some researchers to calibrate the extrinsic parameters more intelligently. Special artificial calibration targets, such as spheres \cite{pereira2016self}, were introduced to avoid manually selecting. However, these methods still shared the limitation of specially designed targets and man-made target errors. To solve this problem, targetless methods were proposed. \cite{taylor2013automatic} was based on the maximization of mutual information obtained between surface intensities measured. In \cite{ishikawa2018LiDAR}, they presented automatic targetless calibration methods based on hand-eye calibration, which only required three-dimensional point clouds and camera images to compute motion information, with a rough calibration result.

In contrast to the above separate approaches for temporal and spatial calibration, some studies focused on integrated calibration. 
Chia-Le Lee et al.\cite{lee2020extrinsic} proposed a method to achieve temporal  and extrinsic calibration between radar and LiDAR based on a specified triangular plate target. And Z. Taylor et al.\cite{taylor2016motion} proposed a targetless self-calibration method that relied on the same motion observed in different sensors to find their relative offsets. 
All these methods prompted us to wonder whether there is an approach to solve the above problems by online integrated temporal and spatial calibration based on environmental features.
Therefore, we propose a method based on the line-based extrinsic calibration method \cite{zhang2021line}, extending the applicability of the extrinsic calibration to temporal and spatial online integrated calibrations.

\section{METHODOLOGY}
In this paper, a sensors pose estimation model is proposed to synchronize sensors data from adjacent moments to the same moment. For the error term equation of the pose estimation model, we adopted the correction based on the environmental line feature to eliminate the influence of the error term on the accuracy, due to the environmental line features are widely existed in indoor and outdoor scenes and has a better correspondence between the camera and LiDAR. Fig.\ref{kuangjia_pic} illustrates the architecture of our method. Our proposed method consists of four steps. 
First, the data between adjacent moments of the camera and LiDAR are projected back to the same moment by means of pose estimation; 
Second, the line features are extracted in a series of ways for both the camera and LiDAR and are filtered and optimized; 
Then, a method for eliminating dynamic point clouds is proposed, which 
only used the association of adjacent two frames instead of detection boxed based on prior information.
Finally, the point cloud line features are projected on the pixel frames, and the score is calculated and optimized during searching optimization periods. Optimizing the score for calibrating the error of the previous stage of the projection of the adjacent frames. The detailed steps are described in the following sections.

\subsection{Pose estimation model}\label{AA}
\begin{figure}
	\includegraphics[width=1.0\columnwidth]{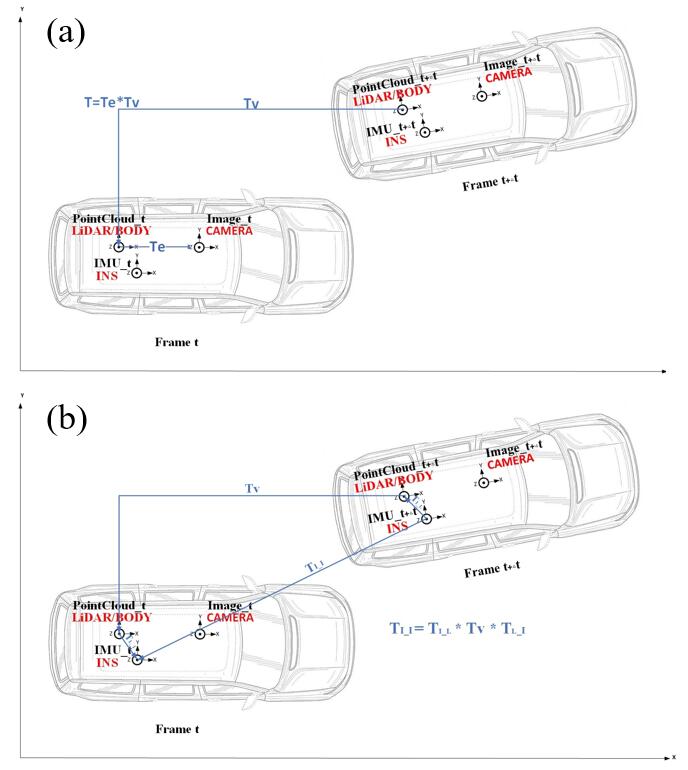}
	\caption{A illustration of pose estimation model. (a) shows the pose estimation model of the camera at time $t$ to the LiDAR at time $t+e_{t}$. (b) shows the motion estimation via the inertial measurement unit navigation conversion.}
	\label{moving_model}
\end{figure}

The temporal and spatial integrated calibration of camera and LiDAR lies in the conversion matrix for temporal and spatial superposition. In this paper, 
we assume that the camera acquires the image at moment $ t $ and the LiDAR acquires the point cloud at moment $ t+e_{t} $ , 
the solution of the conversion matrix between the adjacent moment is solved by the following method:
Estimating sensor motion within $ e_{t} $ period and then
calculating the positional coordinate change between the vehicle body system and sensors system onboard during $ e_{t} $ period by means of coordinate system projection.
According to the pose estimation model:
 \begin{equation}
 \begin{split}
	T&=T_{e}\times T_{v}  \\
	T_{e}&=T_{e}^{\prime}+\epsilon_{e} , T_{v}=T_{v}^{\prime}+\epsilon_{v}
	\label{eq1}
	\end{split}
\end{equation}
where $T$ is the transform matrix from LiDAR at moment $ t + e_{t} $ to camera at moment $t$,
 $ T_{e}$ is the measurement model of LiDAR-camera extrinsic parameters, $ T_{v}$ is the model of LiDAR-camera coordinate system motion estimation as shown in Fig.\ref{moving_model}(a). Since there is an error between measurement and ground truth, 
disassembly of $ T_{e}$ and $ T_{v}$ are shown, then Equation \ref{eq1} is introduced in the next derivation:
\begin{equation}
\begin{split}
	T=\left(T_{e}^{\prime} \cdot T_{v}^{\prime}\right)+\left(T_{e}^{\prime} \epsilon_{v}+T_{v}^{\prime} \epsilon_{e}\right)+\epsilon_{e} \epsilon_{v}
\end{split}
\end{equation}
where $T_{e}^{\prime} \cdot T_{v}^{\prime}$ is measured, $\epsilon_{e} \epsilon_{v} $ is too small to be taken into account, $ T_{e}^{\prime} \epsilon_{v}+T_{v}^{\prime} \epsilon_{e} $ is the error. 

$ T_{e}^{\prime} \epsilon_{v}$ represents the measurement error of the extrinsic parameters, affected by initial values and prolonged vehicle driving.
$T_{v}^{\prime} \epsilon_{e} $ represents the error in the estimation of the LiDAR-camera coordinate system motion. 
The motion of the inertial coordinate system $T_{I_{-} I}$ during $e_{t} $ instant period  is calculated by inertial pre-integration as shown in Fig.\ref{moving_model}(b). Although the uniform linear motion estimation model and the pre-integrated model is quite approximate to the ground truth of the motion of the vehicle body coordinate system , the error term has not to be eliminated since the motion of the LiDAR-camera coordinate system is projected and estimated from the motion of the vehicle body coordinate system, which reduces the accuracy of the projection between adjacent moments $t$ and $t+e_{t}$.
In order to correct the effect of the error term online, line features are introduced in integrated calibration.

\subsection{Line features extraction}
The line features are extracted to correct the error term in the previous stage of optimization. As shown in Fig.\ref{line_features_extraction}, the process of line feature extraction is carried out in both image and point cloud.
For image feature extraction, grayscale processing and canny marginalization are adopted, then line feature extraction algorithm \cite{von2012lsd} is applied.
After that, the model of inverse distance transformation is introduced to deal with the consistent correlation between LiDAR points and image points, which leads to adopting a larger search step to prevent falling into local optimum during searching optimization.

For LiDAR features extraction, the points are divided into different lines, boundary line features are obtained by distance discontinuity.In order to extract sufficient line features on low beam LiDAR,
a local mapping method is applied to combine three frames of point cloud into one, which can present more points in one frame.
Depending on the strength of the GPS signal and the accuracy requirement of the autonomous driving scenario, we propose two local mapping methods: the GPS-based approach and the Normal Distribution Transform (NDT)-based approach\cite{biber2003normal}.
The former method is less computationally intensive with less accuracy, while the latter method is more computationally intensive with more accuracy.

\subsection{Projection and feature filtering}

For matching the line features extracted from the point cloud and the image, pointcloud points set $P^{L}=\left\{p_{1}^{L}, p_{2}^{L}, p_{3}^{L}, \cdots \cdots, p_{n}^{L}\right\}$ and the image points set $P^{C}=\left\{p_{1}^{C}, p_{2}^{C}, p_{3}^{C}, \cdots \cdots, p_{m}^{C}\right\}$ are
created, where $n$ is the number of point cloud points, $m$ is the number of image points, $L$ is the LiDAR coordinate system, and $C$ is the camera coordinate system. We define the rotation matrix $R_{L \rightarrow C}$ and the translation vector $t_{L \rightarrow C}$ , which represent the coordinate transformation process of the point cloud $P^{L}$ projected to $P^{C}$ in the LiDAR coordinate system. 
\begin{equation}
	P_{i}^{C}=R_{L \rightarrow C} \cdot P_{i}^{L}+t_{L \rightarrow C}
\end{equation}

Further filtering is done for outliers in line feature extraction. The above-mentioned LiDAR point cloud has been converted into image form, so a convolution kernel with an 8-pixel boundary is designed to filter out the outliers on the image and get more organized line features from images. In addition, for the points on the ground, are eliminated because their lateral line features do not match well.

\subsection{Dynamic point clouds targets eliminating}
Eliminating the dynamic point cloud targets is extremely essential to the features matching.
In the registration of point clouds at adjacent moments $t$ and $t+e_{t}$, although the stationary targets occupy the dominant position in the observed environment, even a few dynamic targets will bring serious interference to the accuracy of registration, especially in the case of registration based on point cloud line feature rather than the registration based on the original point cloud.

In order to remove dynamic targets, the current widely used method is target detection based on deep learning. Origin point clouds and prior information as the input of the detection network, which could identify possible dynamic objects such as vehicles and pedestrians and return detection box. Then   the point clouds in the detection box are clustered and deleted to eliminate the  dynamic target.
However, this method relies on prior information and an additional detection network, which poses a real-time problem for online integrated calibration.

In order to solve this problem, a lightweight method with excellent accuracy has been proposed.
The K-dimensional binary tree is used to quickly find the nearest neighbors in the multidimensional space by using the dimensional information of the segmentation. The nearest neighbor matching distance for stationary objects is considered to be smaller, while the nearest neighbor matching distance for dynamic objects is larger, so we set a threshold to filter out dynamic targets. Because the point cloud at time $t$ is speculated instead of ground truth of $t+e_{t}$, there should be an angular error. According to the triangle similarity principle, the distant dynamic target needs a larger threshold to be filtered out, so we set a factor to the fixed threshold to make it a linear dynamic threshold. 
Dynamic targets are filtered by linear dynamic thresholds and clustered for subsequent analysis.

\begin{figure}[htbp]
	\includegraphics[width=1.0\columnwidth]{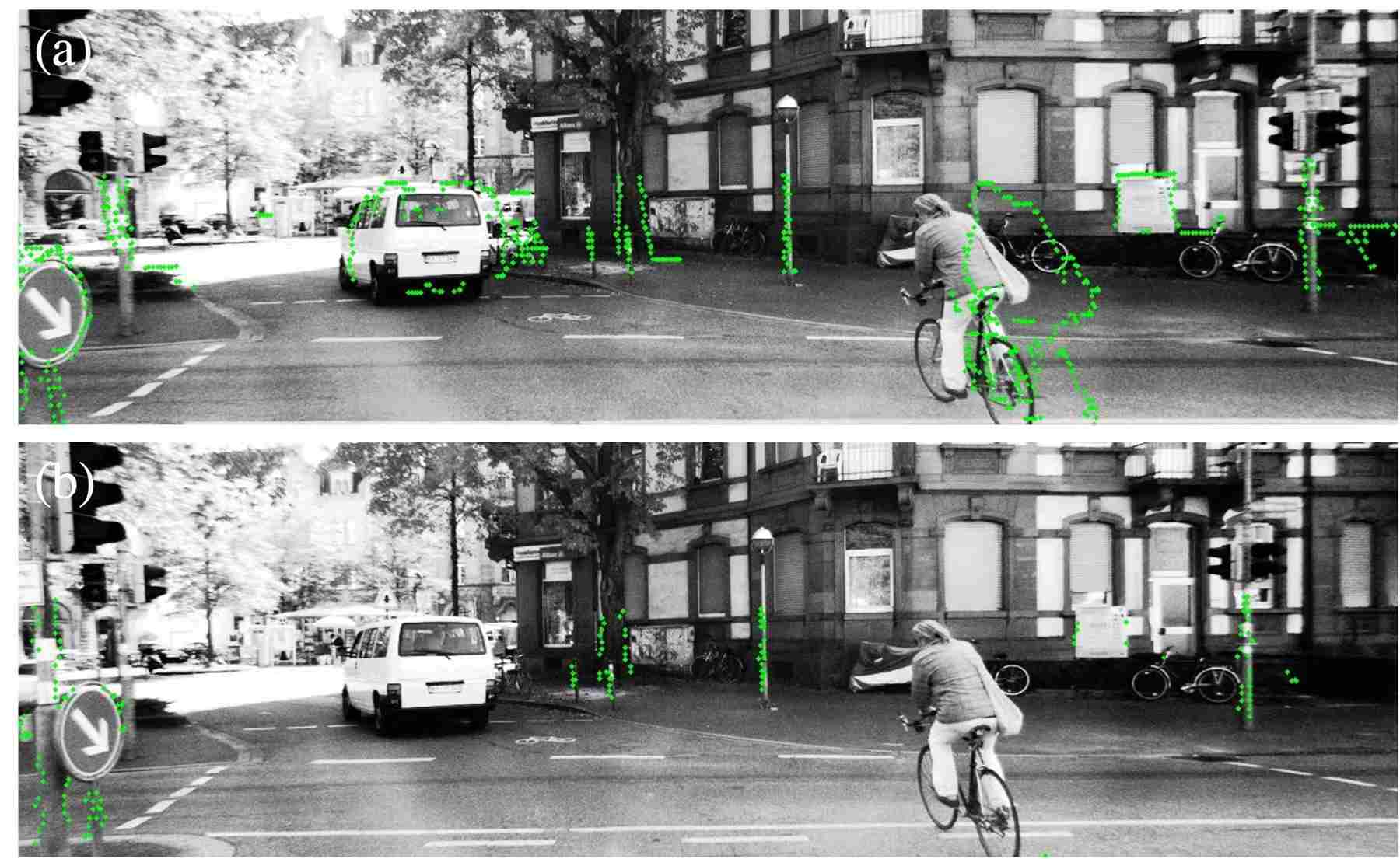}
	\caption{An illustration of how the dynamic targets impact on line feature projection. (a) is the result of  projection without dynamic targets eliminated, and (b) is the result of projection with dynamic targets eliminated. }
	\label{dynamic_erase}
\end{figure}

\begin{algorithm}[thpb]  
 \caption{Optimization process}  
 \label{al2}
 \begin{algorithmic}[1]
 \REQUIRE  
  Image line features $I_t$ at frame $t$, Pointcloud horizontal $F_{h}^{t+e_{t}}$ and vertical $F_{v}^{t+e_{t}}$ line features at frame $t+e_{t}$, Initial extrinsic matrix $T_e$, Motion estimation matrix $T_v$, last frame gray rate $gray\_rate$
\ENSURE Calibrated extrinsic matrix and motion estimation;
  \STATE Initialization: $score$, $max_{score}$ $\leftarrow$ 0.

  \IF{$gray\_rate \textgreater \gamma$}
  \STATE {$step\_size\_larger = \alpha_1$}
  \STATE {$step\_size\_smaller = \alpha_2$}
  \ELSE
  \STATE {$step\_size\_larger = \alpha_3$}
  \STATE {$step\_size\_smaller = \alpha_4$}
  \ENDIF

  \FOR {each $LiDAR\ point$ in $F_{h}^{t+e_{t}}$}
  \STATE $gray\_value = weight * T_e * T_v * p_t$
  \STATE $score += gray\_value$
  \ENDFOR\
  
  \FOR {each $LiDAR\ point$ in $F_{v}^{t+e_{t}}$}
  \STATE $gray\_value = (1-weight) * T_e * T_v * p_t$
  \STATE $score += gray\_value$
  \ENDFOR\
  \IF{$score \textgreater max\_score$}
  \STATE $max\_score = score$
  \STATE Update current\_parameters
  \ENDIF
  \STATE $gray\_rate = score / 255 / points\_num$\\
  \RETURN current\_parameters
 \end{algorithmic}  
\end{algorithm}

\subsection{Searching optimization}\label{op}
In the searching optimization process, an approach with both computation accuracy and efficiency is introduced.
In the previous stage, the LiDAR line features have been extracted and projected onto the line features of the image, and the proportion of LiDAR points projected onto the image gray area has been calculated. 

\begin{figure}[b]
	\includegraphics[width=1.0\columnwidth]{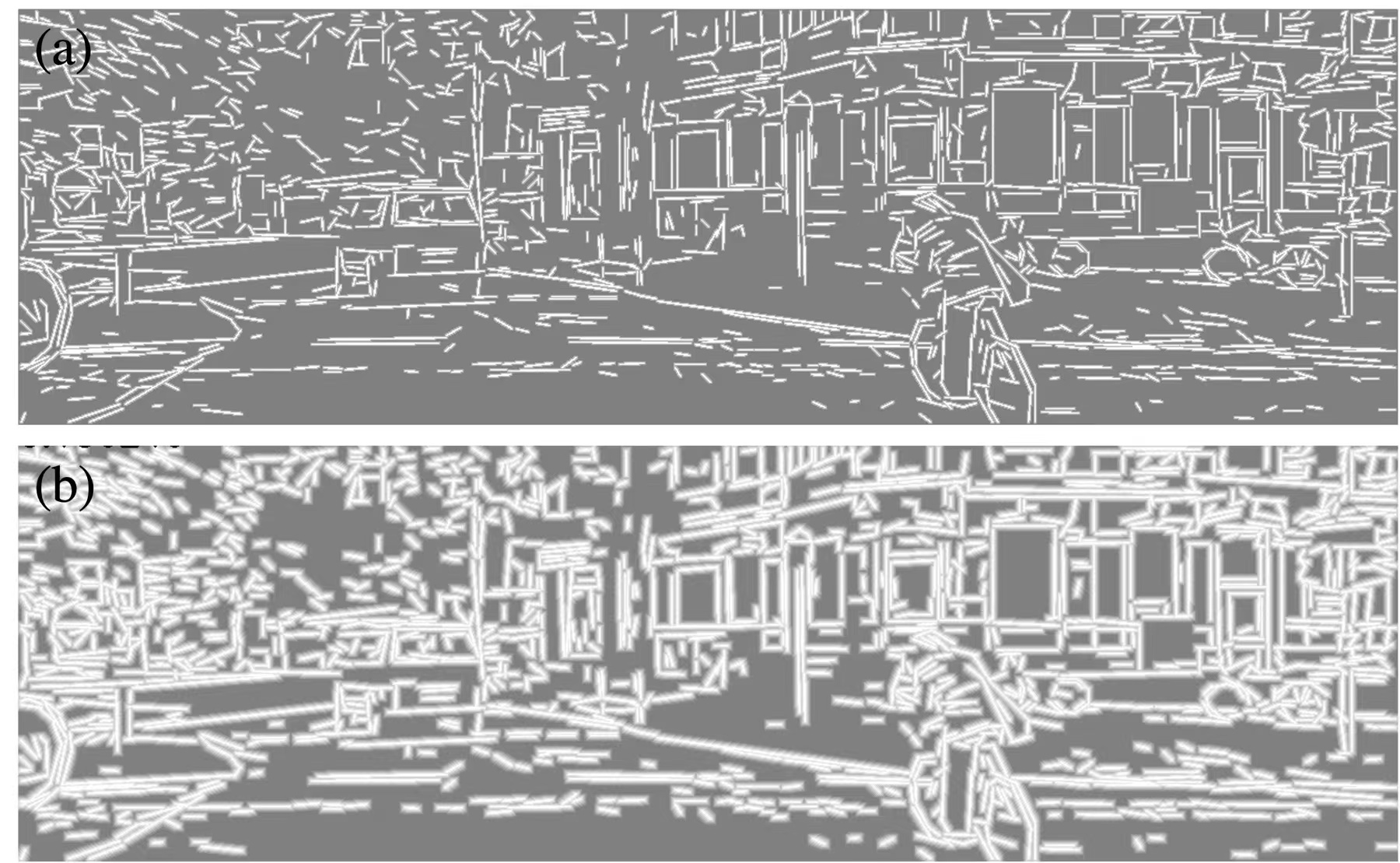}
	\caption{Different gray rate determines different step size strategies. (a) shows a larger gray rate change and (b) shows a smaller gray rate change. }
	\label{filtration}
\end{figure}

For computation accuracy, as is shown in Fig.\ref{filtration}, four searching steps are proposed according to different gray rates.
For computation efficiency, a searching method is applied to optimize the cost function. In \cite{zhang2021line}, they compare the current function score with adjacent 728 scores, if the searching program finds parameters that have a higher score, it will stop the current searching process and begin a new searching process at the position providing a higher score. This searching process will stop when reaching the set iteration count or finding the best score, thus being able to increase the computation efficiency.
We improved this process so that when the direction of the fastest gradient descent is found, we follow that direction to get the fastest optimization.
As shown in Algorithm 1, this is the process of searching optimization.

\section{EXPERIMENT}
In this section, the experiment setups and results of temporal and spatial calibration are described in detail. We validate our proposed approach on KITTI dataset\cite{geiger2013vision}, from which we used a high-resolution camera  and a Velodyne HDL-64E LiDAR. 
The LiDAR was scanned at 10 Hz and the data used in the experiment is synced and rectified. The ground truth temporal and spatial parameters can be obtained from calibration files. 

\subsection{Temporal calibration}\label{tem}

Our method is compared to the approximate strategy of message filter in ROS.  Several evaluation metrics are designed in order to validate the accuracy, which is based on Euclidean distance and ICP and NDT registration.

\begin{itemize}
	\item Euclidean distance-based evaluation metrics: The average Euclidean distance of nearest point in the filter\_cloud is calculated and compared. 
	\item NDT-based evaluation metrics: With the voxel downsampling grid size and other parameters being the same, the final iteration number is calculated and compared.
	\item ICP-based evaluation metrics: with the maximum number of iterations, the maximum distance and other parameters are the same, the score of Euclidean distance is calculated and compared.

\end{itemize}

\begin{table}[htbp]\Huge
	\renewcommand{\arraystretch}{1.2}
	\setlength{\tabcolsep}{1mm}{}
	\caption{CALIBRATION RESULTS OF METRICS}
	\label{table_temporal}
	\centering
	\renewcommand\arraystretch{1}
	\resizebox{\linewidth}{!}{
	\begin{tabular}{lcccr}
		\hline
		\centering
		Project & Dists\_Evaluation\_Score	 & NDT\_Evaluation\_Score & ICP\_Evaluation\_Score
		& \\
		\hline
		\multirow{2}*{\bm{$infer\_cloud\_avg$}} & \multirow{2}*{0.043} & $stepsize1$ 4.091	& $iterationNum1$ 0.042 \\
		~ &  ~ & $stepsize2$ 3.455 & $iterationNum2$ 0.042 \\
		\multirow{2}*{\bm{$infer\_cloud\_interpolation$}} & \multirow{2}*{0.063}  & $stepsize1$	19.10	& $iterationNum1$	0.054\\ ~ &  ~ & $stepsize2$ 6.318 &  $iterationNum2$ 0.044 \\
		\multirow{2}*{\bm{$filter\_cloud\_avg$}} & \multirow{2}*{0.070} & $stepsize1$	33.82	& $iterationNum1$	0.059\\~ &  ~ & $stepsize2$	9.091	& $iterationNum2$	0.045 \\
		
		\hline
	\end{tabular}
	}	
\end{table}

\begin{figure}[bth]
	\centering
	\includegraphics[width=\columnwidth,height=0.24\textheight]{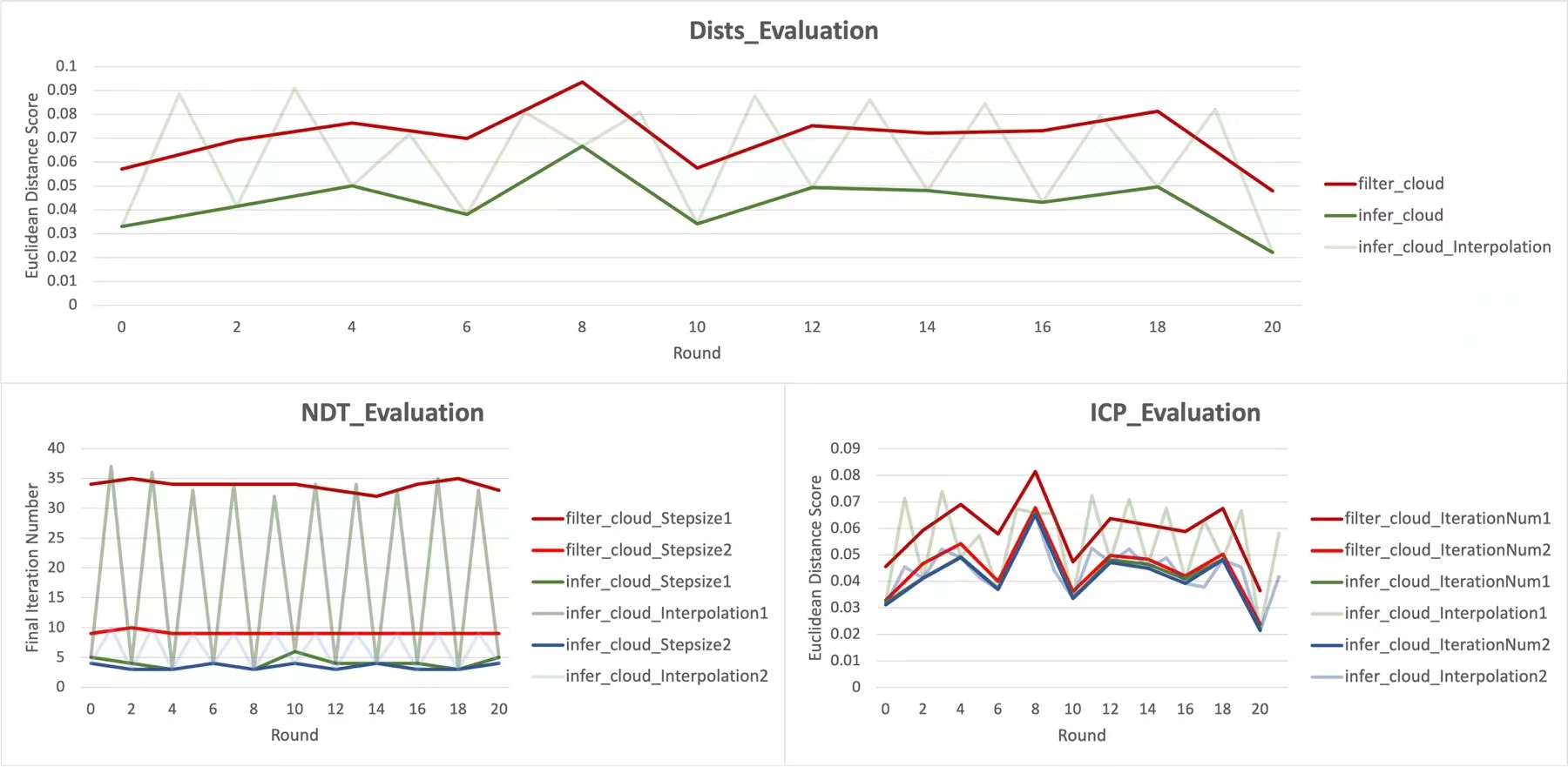}
	\caption{An illustration of three designed metrics performance. (a) shows the Euclidean distance score of Euclidean distance-based evaluation metric; (b) shows the final iteration number of NDT-based evaluation metrics; (c) shows the Euclidean distance score of ICP-based evaluation metrics.}
	\label{experiment_temporal}
\end{figure}

TABLE \ref{table_temporal} shows the score three metrics: $infer\_cloud\_avg$ refers to the average score of a point cloud treated by InferCloud at only 5Hz. $infer\_cloud\_interpolation$ refers to the average score of a point cloud interpolated upward at 10Hz. $filter\_cloud\_avg$ is the average score processed using the ROS soft synchronization method. In NDT evaluation, $stepsize1$ is 0.01m and $stepsize2$ is 0.05m. In ICP evaluation, $iterationNum1$ is once while $iterationNum2$ is five.
The whole results are in Fig.\ref{experiment_temporal}.

The data of Fig.\ref{experiment_temporal} tested based on the scenario where the point cloud frequency is 5Hz, the image and inertial navigation frequency is 10Hz, and the point cloud frame is 0.1s slower than the image. It reflects the advantages of our method and the traditional ROS method in three evaluation metrics.
In Fig.\ref{experiment_temporal}, the red lines ($filter\_cloud$) represent the ROS soft synchronization score curve, and the green and blue lines ($infer\_cloud$) represent the score curve after soft synchronization using our method. Meanwhile, in the NDT/ICP evaluation metrics, we use different step sizes/different iterations to reflect the universality of our method optimization. The green line represents the smaller step length/fewer iterations, and the blue line represents the larger step length/larger iterations.

Since the signal frequency processed by traditional ROS method is determined by the data signal with the lowest frequency, it can be found that the frequency of the $filter\_cloud$ is 5Hz. If only the 5Hz point cloud is processed by our method, and the solid green or blue line is compared with the red line on the image, it can be found that our method is generally superior to the ROS method in all aspects. In addition, our method can also interpolate upward. The shallower line ($infer\_cloud\_interpolation$) below the turquoise is the interpolation score curve. It can be seen that the extra points processed by  interpolation are still with the same level of $filter\_cloud$. Compared with the method of ROS, our approach improves 38.5\% in the Euclidean distance-based metrics and with excellent performance in NDT-based evaluation metrics and ICP-based evaluation metrics.

\subsection{Spatial calibration}\label{extrinsic}
We initially added a 2.0-degrees rotation disturbance on the X, Y, and Z axes to the ground truth parameters. It must be clarified that whether the 2.0-degrees rotation disturbance is positive or negative is stochastic. During the experiment, we compared the calibration error with the ground truth. In addition, we tested the speed of correcting disturbance. 

\begin{figure}[htpb]
	\centering
	\includegraphics[width=\columnwidth]{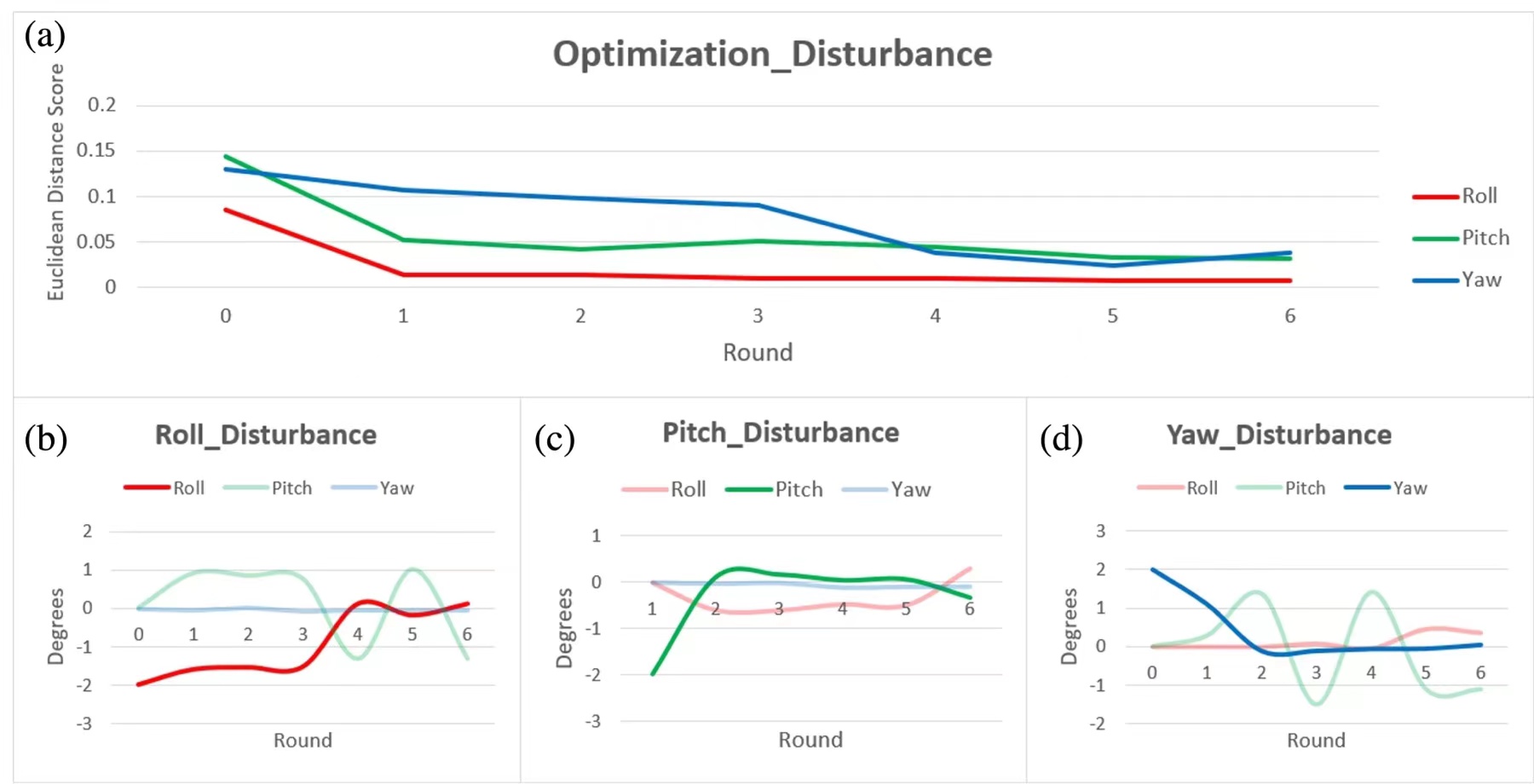}
	\caption{An illustration of instant correction of roll, pitch and yaw disturbances. The red, green and blue lines in (a) represent the correction of disturbance by our method after adding 2.0-degrees disturbance angle in roll, pitch and yaw directions respectively. (b) shows the calibration result of our method when the perturbation angle added 2.0-degrees in the roll direction. (c) shows the calibration result of our method when the perturbation angle added 2.0-degrees in the pitch direction. (d) shows the calibration result of our method when the perturbation angle added 2.0-degrees in the yaw direction.}
	\label{experiment_spatial}
\end{figure}

\begin{figure}[htpb]
	\centering
	\includegraphics[width=\columnwidth,height=0.2\textheight]{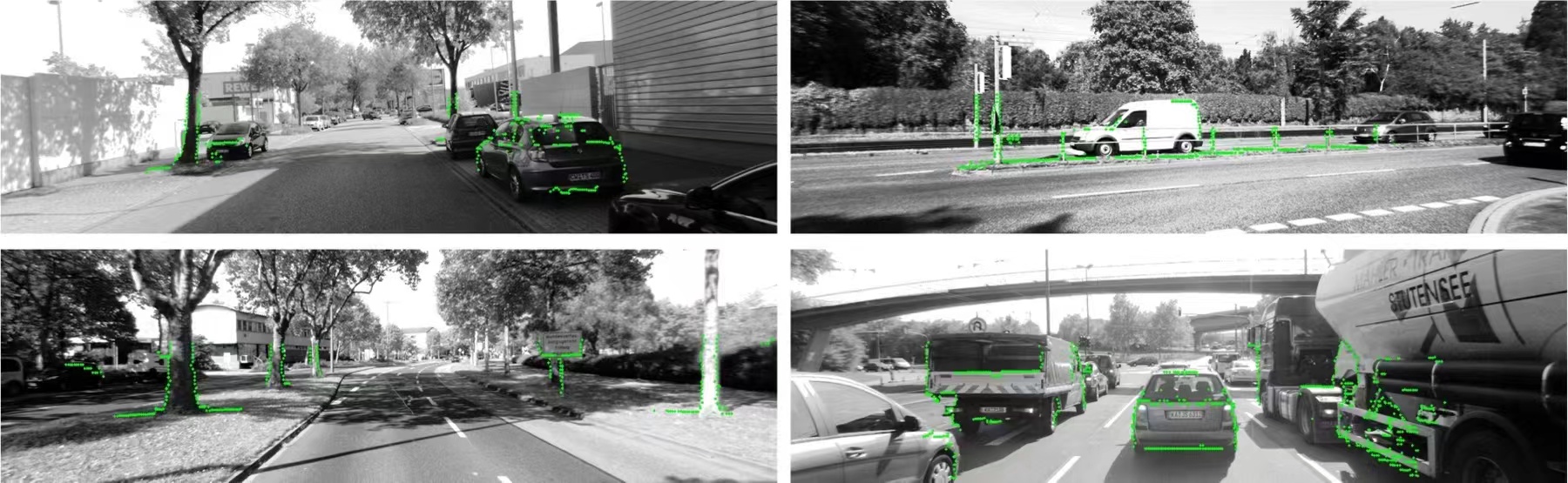}
	\caption{An illustration of spatial calibration of several scenarios. Green points represent the projected LiDAR line points.}
	\label{projection_all}
\end{figure}

It can be seen from Fig.\ref{experiment_spatial} that our method can basically correct the extrinsic parameter drift phenomenon caused by some reasons within 5 frames, and the error generally shows a downward trend and can begin to converge at 1-3 frames. In Fig.\ref{experiment_spatial} (b),(c) and (d), the direction in which the deviation is added is indicated by a darker color. It can be found that from the perspective of the metric of angle correction, our method can also reduce the error and basically complete convergence within 1-3 frames. But there is one small problem that appeared on the direction of pitch angle correction, speculated that for most of us choose picture perspective the middle section of the line on the vertical direction features, and on the pitch direction, if the deviation is also a small angle along with the vertical direction deviation, such as the trunk of a tree after the middle period of line feature extracting, even moving up and down will match the tree trunk well. The overall impact of this oscillation problem is not very big, about $\pm1.0$ degree. And it can be quickly corrected when there are obvious transverse line features. The overall calibration results in more scenarios on the KITTI dataset can be seen in Fig.\ref{projection_all}, which demonstrates that the proposed method is applicable to different scenarios.

\section{CONCLUSION}

The approach for online integrated temporal and spatial calibration for camera and LiDAR is proposed. Based on the pose estimation model of the adjacent moment, line features are selected to correct the error of the model. 
Hardware interface and chessboard are not required in this integrated calibration method. Besides, it's more accurate than the soft simultaneous interpolation method and upward interpolation between the point cloud between adjacent moments is supported. It also demonstrates that the line features of point clouds and images are robust features for matching and calibrating the error of the pose estimation model. 
In addition, we show that the score of the current integrated calibration results can be calculated and further exploited to improve computational efficiency and accuracy. In future work, we plan to introduce temporal and spatial online integrated calibration to multiple cameras and multiple LiDARs.

\section*{Acknowledgment}

This paper and the research behind it would not have  been possible without the exceptional support of atatang Inc.(www.datatang.ai) that kindly provides the High quality on demand professional services specialized in data collection and annotation powering our model. 

\bibliographystyle{ieeetr} 
\bibliography{refs.bib}

\vspace{12pt}
\end{document}